\documentclass[twocolumn,10pt]{asme2e}

\usepackage{graphicx}
\usepackage{xcolor}
\usepackage{url}
\graphicspath{ {./images/} }

\title{SHIP-D: Ship Hull Dataset for Design Optimization using Machine Learning}

\author{Noah J. Bagazinski \thanks{Address all correspondence to this author.}
    \affiliation{
	Department of Mechanical Engineering\\
	Massachusetts Institute of Technology\\
	Cambridge, Massachusetts, 02139\\
    Email: noahbagz@mit.edu
    }	
}

\author{Faez Ahmed
    \affiliation{
	Department of Mechanical Engineering\\
	Massachusetts Institute of Technology\\
	Cambridge, Massachusetts, 02139\\
	Email: faez@mit.edu
    }
}

\begin{document}

\maketitle    
\section*{ABSTRACT}
{\it Machine learning has recently made significant strides in reducing design cycle time for complex products. Ship design, which currently involves years-long cycles and small batch production, could greatly benefit from these advancements. By developing a machine learning tool for ship design that learns from the design of many different types of ships, trade-offs in ship design could be identified and optimized. However, the lack of publicly available ship design datasets currently limits the potential for leveraging machine learning in generalized ship design. To address this gap, this paper presents a large dataset of 30,000 ship hulls, each with design and functional performance information, including parameterization, mesh, point-cloud, and image representations, as well as 32 hydrodynamic drag measures under different operating conditions. The dataset is structured to allow human input and is also designed for computational methods. Additionally, the paper introduces a set of 12 ship hulls from publicly available CAD repositories to showcase the proposed parameterization's ability to accurately reconstruct existing hulls. A surrogate model was developed to predict the 32 wave drag coefficients, which was then implemented in a genetic algorithm case study to reduce the total drag of a hull by 60\% while maintaining the shape of the hull's cross section and the length of the parallel midbody. Our work provides a comprehensive dataset and application examples for other researchers to use in advancing data-driven ship design.}

\section*{INTRODUCTION}
Recent advancements in machine learning for engineering design have shown the ability to create novel designs~\cite{chen2021padgan}, and high-performing systems level designs with significantly reduced cycle time~\cite{mirhoseini2021graph}. The design of ships can greatly benefit from these advancements in machine learning methods as they have years long design cycles and are produced as one-off designs or in small batches. A well-designed machine learning tool for ship design could learn design trade-offs for ships through the continual design of many different types of ships. This can streamline the ship design process, which currently requires large teams of naval architects to balance all the trade-offs in a single ship's design. The current lack of a publicly available dataset for designing ships impinges this possibility. In order to create a machine learning tool capable of generalized ship design, a dataset of ships is needed that represents the vast array of current existing ships. Literature review was unable to find engineering datasets for machine learning for ship design that encompassed the full spectrum of ship shapes needed to generalize the ship design process. The lack of available public datasets is likely due to prior computational limitations, which have now been solved with time as computation has become cheaper and faster. 

This paper presents groundwork for the creation of a dataset of diverse ship hulls to implement machine learning methods for ship hull design. Hull design was chosen as a starting point for the creation of a dataset as this is the traditional starting point in ship design\cite{evans1959basic}. The hull shape affects several key aspects of a ship's performance, including the buoyancy, upright stability, hydrodynamics, and general arrangements of the ship. In addition, the shape of the hull has a direct impact on over 70\% of the cost of a ship~\cite{lin2017feature}. A ship's hull has a significant impact on many aspects of an overall ship system, making it a great candidate to apply machine learning methods to its design to balance overall design trade-offs with a data driven approach.

The following sections detail the literature review of previous work, the methodology for generating a dataset of ships, measures of the dataset, optimization of a ship hull using a trained surrogate model from the dataset, and a discussion on the impact of the work. The dataset of hulls is largely dependent on a parameterization that can represent the broad spectrum of geometric features seen across many traditional hull forms and allows for human and computer inputs to exist together in the same data frame. This parameterization allowed for the creation of a dataset of ship hulls that includes the .stl mesh, images, and hydrodynamic resistance measures of these ship hulls. The key contributions of this paper are:
\begin{enumerate}
    \item Creation of a novel ship hull parameterization to represent a broad spectrum of hull geometries.
    \item Compilation of a set of twelve ship hulls from publicly available CAD repositories to showcase the proposed parameterization's ability accurately reconstruct existing ship hulls, which can be used as a benchmark for future studies in generalized ship hull design representation.
    \item A publicly available dataset of ship hulls to implement data driven approaches in ship design. Each hull in the dataset has the parametric representation, meshes, images, and hydrodynamics drag measurements. 
    \item A case study demonstrating surrogate based optimization using a residual neural network and genetic algorithms for ship hulls.   
\end{enumerate}

\section*{PREVIOUS WORK}
This section reviews previous work that informed the work presented in the remainder of the paper. The first subsection provides a background in generating datasets for engineering problems, providing inference to the size, scope, and contents of good engineering datasets. The second subsection investigates prior work in ship hull design representation, showing that a single design representation for the diversity of ships needs to be created. The third subsection reviews methods for machine learning for hull design, showcasing that current work in the field focuses on hydrodynamic optimization. To enable the current practices of the field with this dataset, the final subsection overviews different methods of predicting the hydrodynamic resistance, or total drag, of ships, showcasing that linear potential flow solvers for predicting wave drag balance accuracy and computational efficiency for use in dataset generation. 

\subsection*{Dataset Generation for Engineering Design}
The creation of a dataset is paramount for data driven design and surrogate modeling of a design's performance. The two critical components of a dataset for engineering design are a design representation and performance metrics for each sample in the dataset. Publicly available engineering datasets found in a literature search include bicycles~\cite{regenwetter2022biked,regenwetter2022framed}, linkage systems~\cite{heyrani2022links}, metamaterials~\cite{chan2021metaset,lee2023tMetaset}, and ships~\cite{read2009drag}. The subsection on design representation will continue to explore this dataset of ships as well. Other work has shown that multi-modal information can lead to improved accuracy in training a machine learning model~\cite{song2023attention}. In order to provide the best possible results for future work for machine learning for ship design, multi-modal information on the representation, shape, and performance will be included in this dataset. Sample size among these public datasets range from several hundred~\cite{read2009drag} to over a hundred million~\cite{heyrani2022links}. In order to create the most impact for the design of ship hulls with machine learning techniques, a dataset that can encompass most traditional ship hull designs will need to be construed. The goal of the number of samples in this paper's dataset is to provide broad coverage of the entire feasible domain of ship hulls. An analysis on this is provided in the Results and Discussion Sections. 

\subsection*{Design Representation}
A dataset of ship hulls will need a design representation that is comprehensive enough to cover the broad spectrum of traditional ship hull forms. A literature review has shown multiple representations for complex designs, including graphs~\cite{heyrani2022links, mirhoseini2021graph}, images~\cite{maze2022topodiff,lee2023tMetaset}, parameterized vectors~\cite{chan2021metaset,chen2021padgan, brown2003multiple, HullOpt, read2009drag, khan2022shape, khan2022geometric, zhang2018parametric, chrismianto2014parametric, lu2016hydrodynamic, PSOShip_Opt,PSO_Multi_Opt}, and free form deformation techniques~\cite{wang2022shipEncoding,ao2021artificial,ao2022artificial,peri2001design,demo2021hull}. The most common representation found for ship hulls was vectored parameterization. Many of these human defined parameterizations allow human users to create their own designs with few ($<$10) inputs. This created a lack of diversity among the geometric features found in the design space of these parameterized representations. Meanwhile, free-form deformation techniques allow the greatest diversity in geometry, but they require an initial design to seed the deformation process, limiting shape diversity.  In order to allow for reasonable human input in ship hull design, a parameterization with a broad definition of geometric features will need to be developed to represent a large design space that encompasses most traditional hulls. Dimensionality analysis performed on different parent hulls by Wang et al. and Kahn et al found that 32 and 27 learned parameters can reasonably reconstruct complex surface features on hulls, respectively. As these two sources only analyzed the dimensionality of single hull forms, it is likely that the hull parameterization's dimensionality will require more than 32 dimensions to cover the desired diversity, although it is likely within a similar order of magnitude. The Methods Section will show that the dimensionality of the developed parameterization for a diverse spread of hull forms is 45 parameters. 

\subsection*{Machine Learning for Engineering Design}
The goal of releasing the dataset produced in this paper is to provide it for machine learning researchers train their models for the improvement in data driven hull design. Outside the realm of ship hulls, machine learning methods applied for design purposes include classification~\cite{li2022machine}, reinforcement learning~\cite{mirhoseini2021graph,li2022machine}, data augmentation~\cite{maze2022topodiff,chen2021padgan}, and surrogate modeling for optimization~\cite{maze2022topodiff,li2022machine,chen2021padgan}. Within applications for ship hull design, machine learning practices have primarily focused on surrogate modeling of the hull's hydrodynamics~\cite{wang2022shipEncoding,ao2021artificial,ao2022artificial,khan2022geometric,khan2022shape,read2009drag,marlantes2021Modeling}. Additional work in machine learning for ship hulls noted in the previous subsection is dimensionality analysis of geometric features on ship hulls. In order to continue the prominent current direction of research in machine learning for hull design, hydrodynamic measures of hulls will be captured in the dataset to enable surrogate modeling of a hull's hydrodynamics. The next subsection details the different methods of measuring total drag to weigh computational effort versus simulation accuracy. 

\subsection*{Hydrodynamic Resistance Prediction}
Several methods for measuring the hydrodynamic resistance of ship hulls were considered. This section details these methods. 

Traditionally, before the advent of computational tools, the resistance of a ship hull was measured in a towing tank with a scaled model of the hull. Guidance by the International Towing Tank Conference (ITTC) gives the process of scaling the total drag of a model hull into total drag estimate of a full sized ship~\cite{AppNavArch}. Measuring the drag of with a scaled physical model test is the most accurate method of predicting the drag of a full sized hull as the hull is measured in real water as opposed to a simulation's model for water and fluid dynamics. However, this method is too time and cost intensive to produce a dataset for machine learning, since each individual hull would need to be constructed for testing. Towing tank tests, however, are used to benchmark computational predictions of total drag~\cite{noblesse2013neumann,noblesse1983proceedings,huang2013numerical, newman2018marine, yang2013practical}. Included in this paper are the parameterized reconstruction of two of these benchmark hulls: The Wigley Hull~\cite{noblesse1983proceedings}, and the DTMB 5415 Hull~\cite{olivieri2001towing}. Computational models of ship drag, on the other hand, provide cost effective and accurate measures of drag. 

Computational models vary in computational effort and accuracy. Computational fluid dynamics (CFD) solvers are commonly used to create accurate simulations of drag on a ship at the expense of high computation time. CFD solvers have been used in several prior works for design optimization~\cite{chrismianto2014parametric,demo2021hull} and machine learning for ship design~\cite{wang2022shipEncoding}. Other predictions of drag rely on empirical predictions of drag such as Savitsky's method~\cite{savitsky1964hydro} and Hollenbach's method~\cite{hollenbach1998estimating,hollenbach2007efficient}. These computational models are not good for a dataset generation as these models are limited to specific types of ships These models take geometric measures as inputs, but not the 3D model of the hull itself, limiting the scope of applicability for dataset generation. Many papers used empirical regression models in ship design optimization to arrive at principle dimensions of a hull~\cite{hart2010IMDO,diez2010robust,brown2003multiple, HybridAgent, PSOShip_Opt,PSO_Multi_Opt} and in machine learning applications to improve performance prediction~\cite{marlantes2021Modeling}. 

Another simulation method, linear wave solvers, provide accurate results to drag measurement with reduced computational effort relative to CFD. These solvers use potential flow to simulate the waves produced by a ship in steady forward motion to estimate drag from propagating waves. Different linear wave solvers including Michell's Integral~\cite{michell1898Wave,tuck1989wave}, Rankine Panel Methods~\cite{mantzaris1998rankine}, Neumann-Kelvin Theory (also called Dawson's Method)~\cite{dawson1977practical}, and Neumann-Michell Theory~\cite{noblesse2013neumann,yang2013practical, huang2013numerical}. These potential flow solvers input the 3D geometry of a hull and provide accurate measures of drag at typical operating speeds of a hull. As seen in the literature review, the balance of computational speed and accurate results make potential flow solvers great candidates for both optimization and machine learning methods for ship hull design~\cite{ao2021artificial, ao2022artificial, khan2022geometric, khan2022shape, peri2001design, read2009drag, lu2016hydrodynamic, HullOpt}
Among the available solvers, the Michell Integral was chosen as the simulation for hulls in this dataset.

\section*{METHODS}
This section details methods used to define the hull parameterization for the dataset, validate the parameterization's ability to construct a diversity of ship forms, and train the surrogate model for drag prediction.

\subsection*{Hull Parameterization}
This section details the hull parameterization and methods for generating aspects of the data set, such as the meshes and images of each hull.

\subsubsection*{Parameterization Terms}
The proposed parameterization encompasses broad features seen in traditional ship hull geometries. As mentioned in the prior work section, prior parameterizations used for ship hull analysis characterized ship hulls with traditional measures of ships hulls, such as block coefficient, midship coefficient, and waterplane coefficient. While these traditional characteristics can allow for the rapid generation of some geometric aspects of a hull, they cannot fully represent a diversity of hull forms, nor do they contain  enough information to generate a final hull form. The proposed parameterization characterizes geometric features found on traditional hull forms using measures of angles and length ratios, which are applied to a set of algebraic equations to define points on the hull's surface. 

The proposed hull parameterization is made up of 45 terms. These terms were construed through analyzing and characterizing the shape and curvature of many different publicly available hull geometries. Some of these hulls were chosen to be a part of the set of target hulls seen in Figure~\ref{fig:TargetHulls}. The following breakdown of the parameters also follows the process of first generalizing the shape of the hull, breaking down the ship into sections, and defining specific parameters that encompass the geometric features seen in each section. The first seven terms define the main principal dimensions of the hull. These terms include the length overall, the beam at the main deck, the beam at the stern, and the depth of the hull. The next four terms define the cross section of the parallel midbody of the hull. The cross section terms are the deadrise angle, the chine radius, the keel radius, and the beam of the chine. Twenty terms define the geometry of the bow and stern taper of the hull. These terms characterize the shape of the bow and stern rake, the keelrise, the transition from the taper to the parallel midbody, the drift angle from the bow across the hull's depth, and the cross section of the transom. The final fourteen terms define the geometry of bulbs at the bow and stern, which were inspired by parameterizations of bulbous bows found in literature~\cite{chrismianto2014parametric, zhang2018parametric}. These parameters characterize the dimensions, vertical asymmetry, and fillet to transition the bulb into the hull. Overall, these 45 terms are characterized using a human understanding of ship hull geometry and are labeled to allow for human input, in addition to having a vectored structure for computer generated input as well. These 45 terms are intended to characterize a diversity of curvature and shapes seen across large ships to small recreational boat hulls, so that the design of most hulls can be all characterized in the same design representation. A study on the accuracy of reconstructing existing hulls is described later in the Methods Section. 

These parameters populate a set of algebraic equations that define the surface of the hull. By characterizing the shape of the hull with a set of equations, the hull can be characterized and measured at any fidelity, which allows for a large range of computational opportunities to characterize and measure the hulls. The section on meshing later in the Methods section details the construction of the surface of a hull. 

\begin{figure}[ht]
\begin{center}
\setlength{\unitlength}{0.012500in}%
\includegraphics[width = \columnwidth]{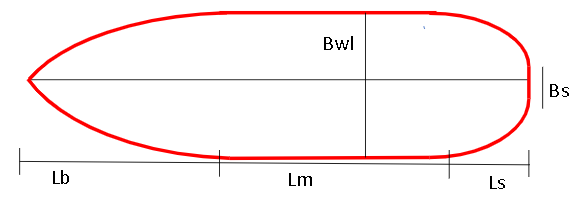}
\caption{Seven terms define the principal dimensions of the hull, including the length, beam, depth, draft, and tapers at the ends of the hull.}
\label{figure_PARAM1} 
\end{center}
\end{figure}
\begin{figure}[ht]
\begin{center}
\setlength{\unitlength}{0.012500in}%
\includegraphics[width = \columnwidth]{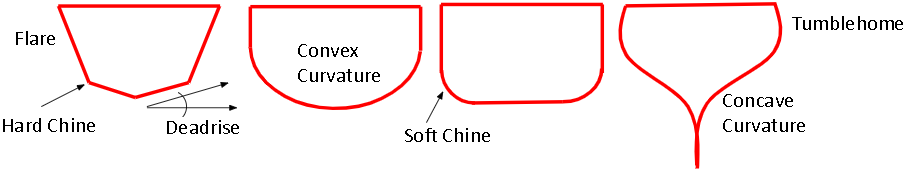}
\caption{Four terms define the cross section of the hull in the parallel midbody. These terms can create cross sections seen on traditional hulls ranging from chines, bilges, flare, tumblehome, and S-chines.}
\label{figure_PARAM2} 
\end{center}
\end{figure}
\begin{figure}[ht]
\begin{center}
\setlength{\unitlength}{0.012500in}%
\includegraphics[width = \columnwidth]{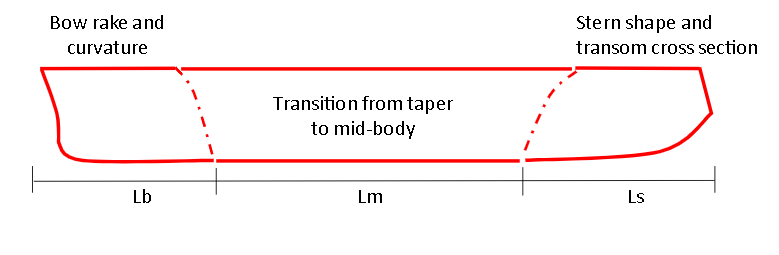}
\caption{Twenty terms define the tapered regions at the bow and stern of the hull. These terms define features such as drift angle, keelrise, transom cross section, rake, and the transition from the taper to the parallel midbody.}
\label{figure_PARAM3} 
\end{center}
\end{figure}
\begin{figure}[ht]
\begin{center}
\setlength{\unitlength}{0.012500in}%
\includegraphics[width = \columnwidth]{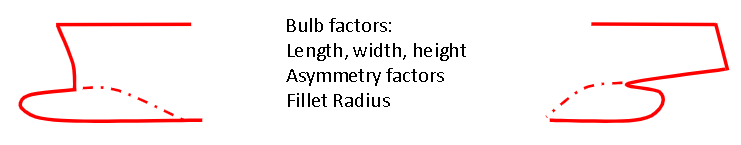}
\caption{Fourteen terms define the bow and stern bulb geometries, including terms that define the size, vertical asymmetry, and the fillet transition of the bulb into the hull.}
\label{figure_PARAM4} 
\end{center}
\end{figure}

\subsubsection*{Constraint Definitions}
While the parameterization can define a large design space of hull geometries, constraints on the parameterization are needed to ensure that a feasible hull will be produced by a specific set of parameters. To satisfy a ``feasible'' hull shape, the hull's surface only needs to satisfy two criteria: 
\begin{enumerate}
    \item The hull is watertight, meaning that there are no holes in the surface of the hull.
    \item The hull surface is not self-intersecting.     
\end{enumerate}
As the hull surface is defined by a set of equations with constants dictated by the parameter values, conditions to determine whether a hull's surface satisfies the two main feasibility criteria can be solved algebraically. The advantage to algebraically solving these conditions is significantly reduced computational effort to algebraically check hull feasibility compared to feasibility checks with mesh generation. After searching through the design space of the hull parameterization and examining the equations that define the hull surface, a set of 49 constraints were defined to determine if a hull surface produced from a specific parameterization satisfies the two feasibility criteria. Mesh generation and feasibility checks are computed in $O(Nlog(N))$, where $N$ is the number of vertices on the mesh. For comparison, on an Intel Core i9-10980XE processor, the construction and check of a hull mesh with approximately 80000 vertices is 1.77 seconds, while the algebraic constraints check feasibility in 0.000199 seconds. This is a $10^4$ increase in speed for checking hull feasibility with the algebraic constraints.  

\subsubsection*{Surface Generation and Meshing}
In conjunction with the parameterization terms, a set of equations was developed to generate the surface of the hull. Terms expressing the cross section define a set of lines that are tangent to circular curves to create the keel and chine. With this, terms for the bow shape, drift angle, and taper endpoints for a given waterline height give four boundary conditions to define a cubic polynomial to define the $(X,Y)$ points along the bow curvature for a given $Z$ position. This is similarly true for the stern taper, although there is extra consideration for the transom cross section. Additionally, terms for the bulbs define ellipsoid surfaces that can be controlled with the parameterization terms. The bubs merge to the hull via fourth order polynomials to fillet the bulb to the remainder of the hull. 

With the set of equations, any point cloud with custom spacing of $X,Y,$ and $Z$ can define the hull's surface. The meshes of the hulls provided in the data set were constructed from these point clouds of the hulls with even spacing between the X and Z coordinates. Further provided with the dataset are five images of each hull mesh:
\begin{enumerate}
    \item Front View
    \item Profile View
    \item Plan View
    \item Three-Quarter Starboard Bow View
    \item Three-Quarter Port Stern View
\end{enumerate}

\subsection*{Dataset Generation}
The following section details the generation of the hull parameterizations in the dataset. The hulls generated in the dataset were randomly generated and made up of three distinct subsets of hulls. Each term in the parameterization was sampled uniformly from its range of possible values.  Many of the terms in the parameterization are relational and have limits between 0 and 1. Other terms rely on user defined inputs to ensure that the generated designs are similar to realistic ship hulls. For example, the term related to the beam-length ratio of the hull was limited to be between 0.0833 and 0.333 to ensure that the beam-length ratio of the dataset hulls encompasses that of typical ship hulls. This is similarly true for the term related to the depth-length ratio, which was limited to be between 0.05 and 0.25. Additionally, the term related to the deadrise angle of the cross section was limited to be between $0^{\circ}$ and $45^{\circ}$. After generating a random parameterization, the forty nine algebraic constraints were checked for each given random parameterization. If the randomly generated parameterization led to a feasible hull shape, then it was added to the dataset. This process was repeated until each subset contained ten thousand hulls, for a total of thirty thousand hulls in the dataset. The purpose of randomly generating the parameterized hulls was to create a dataset that fully encompasses the possible design space of ship hulls that meet the feasibility criteria so that a machine learning model can learn the relative performance of a hull's geometric features in isolation and in combination. The training of a machine learning model to predict the drag coefficient of hulls is detailed in a later section of the paper. 

The dataset is comprised of three subsets of ten thousand hulls. The first subset of the dataset contains hull forms generated from the full possible range of each term in the parameterization. These hulls contain all the possible combinations of all the geometric features defined by the hull parameterization. The second subset of hulls is constrained so that they do not contain bulbs. By allowing the full range of all the terms except those that define bulbs, geometric features that are typically seen in smaller hull forms are more prominent in this subset. Smaller hull forms include hulls that are typically less than fifty meters in length, such as tugboats, fishing trawlers, ferries, yachts, and recreational watercraft. Such geometric features include deadrise, concave cross sections, and chines. Meanwhile, the third subset contains hulls that are biased towards features seen in larger hulls. These hulls have a keel radius that is strictly positive and have a zero degree deadrise angle. By eliminating these features and allowing for the presence or absence of bulbs, this subset contains hulls with geometric features that are more prominent in larger hull forms. Larger hull forms include those seen on warships, cargo ships, cruise ships, and research vessels. The slight biases introduced within the latter two subsets ensures that there exist samples with features akin to realistic hulls.  This will give surrogate models trained on the whole dataset more information to accurately predict the performance of realistic hulls. This will ensure surrogate models trained on this dataset can yield reasonable predictions for any hull and accurate predictions for realistic hulls. 

\subsection*{Chamfer Distance Comparison of Hulls}
The proposed parameterization is intended to represent a large diversity of hull forms. However, how does one validate if a proposed parameterization is sufficiently expressive? One solution is to collect a diverse set of real-world hull forms and check if the proposed parameterization is capable of recreating those hulls. This strategy  is used to validate the usefulness of the proposed parameterization. The proposed method gathers a small set of realistic hulls and generates their surfaces as point clouds. Then, custom parameterized hulls are also constructed to match the point cloud of the target hulls with point clouds generated by the hull parameterization. The set of twelve target hulls listed in Table ~\ref{tab:MatchOpt}.

The first ten hulls were gathered from GrabCAD, an online repository of 3D CAD models. The final two, the Wigley Hull and the DTMB 5415 Hull, are hulls commonly used as benchmarks in hydrodynamics computation and tow tank testing\cite{newman2018marine,noblesse1983proceedings,noblesse2013neumann,tuck1989wave,olivieri2001towing,huang2013numerical}. These hulls represent a large diversity of ship hull shapes and scales, including recreational watercraft, commercial ships, and warships.  Parameterized reconstruction of these twelve hulls is detailed in the Results and Discussion Sections. 

The metric used to evaluate the match between two point clouds is the bidirectional mean of squared Chamfer distances. For two point clouds, $A$ and $B$, the Chamfer distance finds the distances from each point in $A$ to its nearest neighbor in $B$. The distance metric used is the squared Euclidean distance between two points. Bidirectional Chamfer distance is calculated for all points in $A$ to $B$ and for all points in $B$ to $A$. The sum of all the square distances is then averaged to form the bidirectional mean of squared Chamfer distances. The formula for this evaluation metric is shown below:

\begin{equation}
CD = \frac{1}{N_A + N_B} ( \sum_{n=1}^{N_A} ||A_n - B_{n*}||^2 +
\sum_{n=1}^{N_B} ||B_n - A_{n*}||^2 )
\label{eq_CD}
\end{equation}
where $B_{n*}$ is the nearest neighbor of the point $A_n$ in $B$ and $A_{n*}$ is the nearest neighbor to the point $B_n$ in $A$. $N_A$ and $N_B$ are the total number of points in $A$ and $B$, respectively.
While the parameterization is designed to be manipulated by a human designer, it is difficult to manipulate the parameterization by hand to match another hull. In order to reconstruct the set of target hulls as parameterized hulls, a genetic algorithm was written to minimize the bidirectional mean of the square of Chamfer distances. The reconstructed target hulls generated from the optimization are shown in the Results Section.  

\begin{figure*}[ht]
\begin{center}
\setlength{\unitlength}{0.012500in}%
\includegraphics[width = \linewidth]{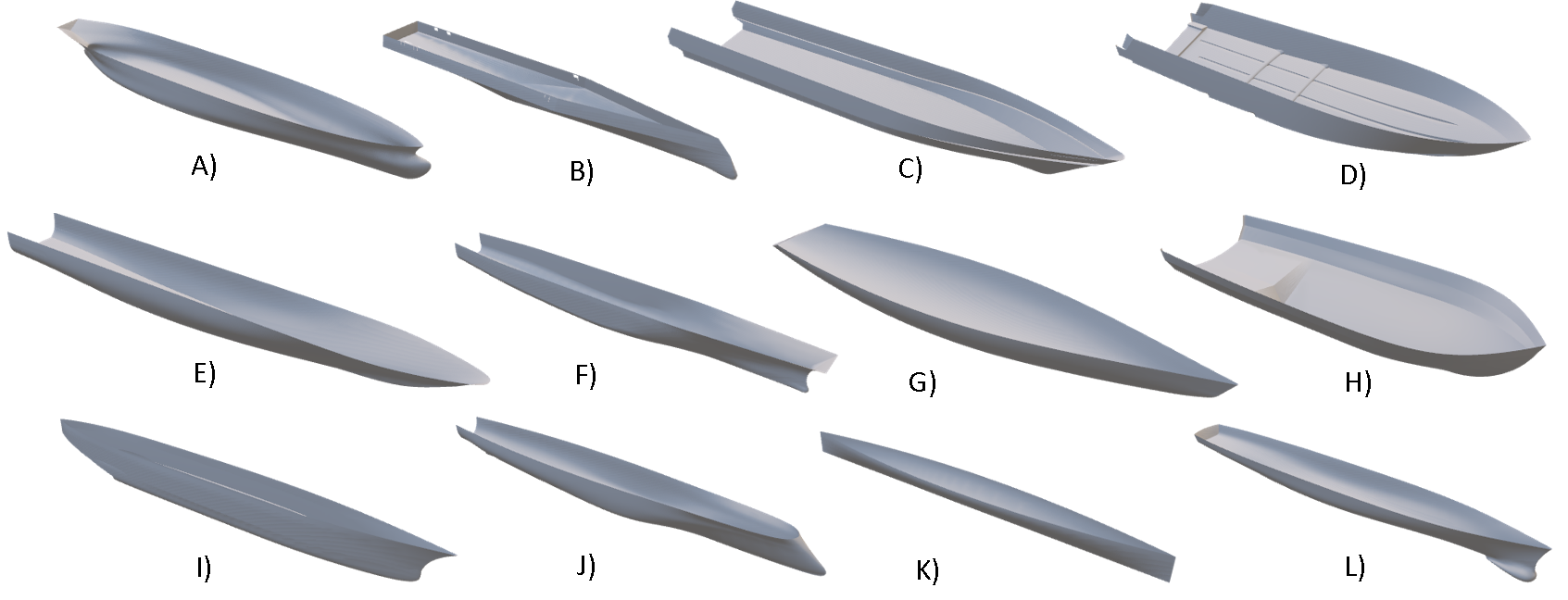}
\caption{The set of twelve target hulls used for validating the capability of the hull parameterization to reconstruct a diverse set of realistic hull forms.}
\label{fig:TargetHulls} 
\end{center}
\end{figure*}

\subsection*{Hull Resistance Calculation}
As noted in the Background Section, the method to simulate the total drag of each hull in the dataset will be a linear wave solver to simulate the wave drag and  with the ITTC regression line to predict viscous drag on the hull. The following subsections detail the method of calculating the wave drag, validating the simulation, and training a surrogate model to predict the wave drag coefficient of a hull from its parameterization. 

\subsubsection*{Michell Integral}
The Michell Integral was chosen to simulate wave drag over other linear wave methods for its relative computational efficiency for the accuracy it provides. The Michell integral is a linear estimate of wave drag of a slender ship in forward motion. The model performs a Fourier series analysis on the waves that propagate from the hull and thus, this model is not time dependent, leading to its computational efficiency. The Michell Integral is defined by the following equation~\cite{michell1898Wave,tuck1989wave}: 

\begin{equation}
 R_w = \frac{A \rho g^2}{\pi U^2} \int_{1}^{\infty} (I^2 + J^2)* \frac{\lambda^2}{\sqrt{\lambda^2 - 1}} d\lambda
\label{eq_Mich}
\end{equation}
 where $\rho$ is the density of water, $g$ is gravity, $U$ is the ship speed, and $A, I, and J$ are integrated terms relating to the surface normal across the hull and the direction of wave propagation. Further insight into these terms is in Michell's paper form 1898~\cite{michell1898Wave}.
 
 Using the Michell Integral, thirty two wave drag coefficients were calculated for each hull across four different draft and speed operating conditions. The four drafts were 25\%, 33\%, 50\%, and 67\% of the hull's total depth. The eight speed conditions were normalized to Froude numbers between 0.15 through 0.45 in steps of 0.05. These Froude numbers correspond to typical operating conditions of traditional displacement hulls~\cite{AppNavArch,newman2018marine}. The Froude number is the relative scaling between inertial and gravitational forces described in the equation below:
\begin{equation}
F_n = \frac{U}{\sqrt{gL}}
\label{eq_Fn}
\end{equation}
 Where $U$ is the hull speed, $g$ is gravity and $L$ is a length scale. The length used in simulating the 32 speed-draft  conditions of the hulls was the length of the waterline at the tested draft mark. This way, thirty two unique conditions were measured. As the wave drag is a function of the hull geometry and the interference a propagating wave makes with the hull, a full spectrum of speed and draft marks were calculated for the dataset. For the purposes of applying machine learning to this dataset, including a full spectrum of speed-drag conditions in the dataset allows a machine learning model to predict the drag at multiple operating conditions as opposed to only one operating condition. Providing all this information allows the model to learn the effects of drag due to changing submerged geometry with draft and speed. In addition to scaling the relative speed and draft conditions for the hulls, the wave drag is also scaled using the following equation:
 \begin{equation}
  C_w = \frac{R_w}{\frac{1}{2} \rho U^2 LOA^2}
\label{eq_Cw}
\end{equation}   
Typical drag coefficients of hulls are scaled by the wetted surface area of the hull. Within the dataset, however, the wetted surface area of the hulls can vary greatly. Instead, the Length-Overall (LOA) is used instead as this is the first term in the parameterization. For the purposes of machine learning with the dataset, the wave drag coefficient can be characterized by the remaining 44 terms in the parameterization and the hull's relative speed and draft. With the thirty two wave drag coefficients, any speed-draft condition within the range of the dataset conditions can be interpolated. The calculation of the thirty two wave drag coefficients for the thirty thousand hulls in the dataset was performed in parallel on an Intel Core i9-10980XE processor. The average computation time for an individual hull was 72 seconds for the thirty two wave drag coefficients. 
 
\subsubsection*{Total Resistance}
For displacement hulls, the principle characteristics of drag are defined as the sum of residual drag and skin friction drag. Skin friction is viscous drag due to boundary layer effects of water across the surface of the hull. In traditional naval architecture, the skin friction drag is approximated from a series of regression tests performed by the ITTC\cite{AppNavArch}. The skin friction coefficient regression is:
 \begin{equation}
C_f = \frac{0.075}{(Log_{10}(Re) - 2)^2}
\label{eq_Cf}
\end{equation} 
where $Re$ is the Reynolds number of the hull, scaled with its forward velocity, and length. As the skin friction coefficient scales with the wetted surface area of the hull, the total skin friction drag scales with:
 \begin{equation}
R_f = \frac{1}{2}C_f\rho U^2 A_{ws}
\label{eq_Rf}
\end{equation}
where $R_f$ is the skin friction resistance and $A_{ws}$ is the wetted surface are of the hull. The other component of ship drag, residual drag, is the sum of viscous pressure drag and wave drag. As ships' hulls are considered slender, the contributions of viscous pressure drag are negligible relative to the scale of wave drag. With this consideration, the total Resistance, $R_t$ is the sum of wave drag and skin friction drag:
 \begin{equation}
R_t = R_w + R_f
\label{eq_Rt}
\end{equation}
The following subsection details the validation of this assumption with the DTMB 5415 hull form. 
\subsubsection*{Wave Drag Validation}
\begin{figure}[t]
\begin{center}
\setlength{\unitlength}{0.012500in}%
\includegraphics[width = 3.5in]{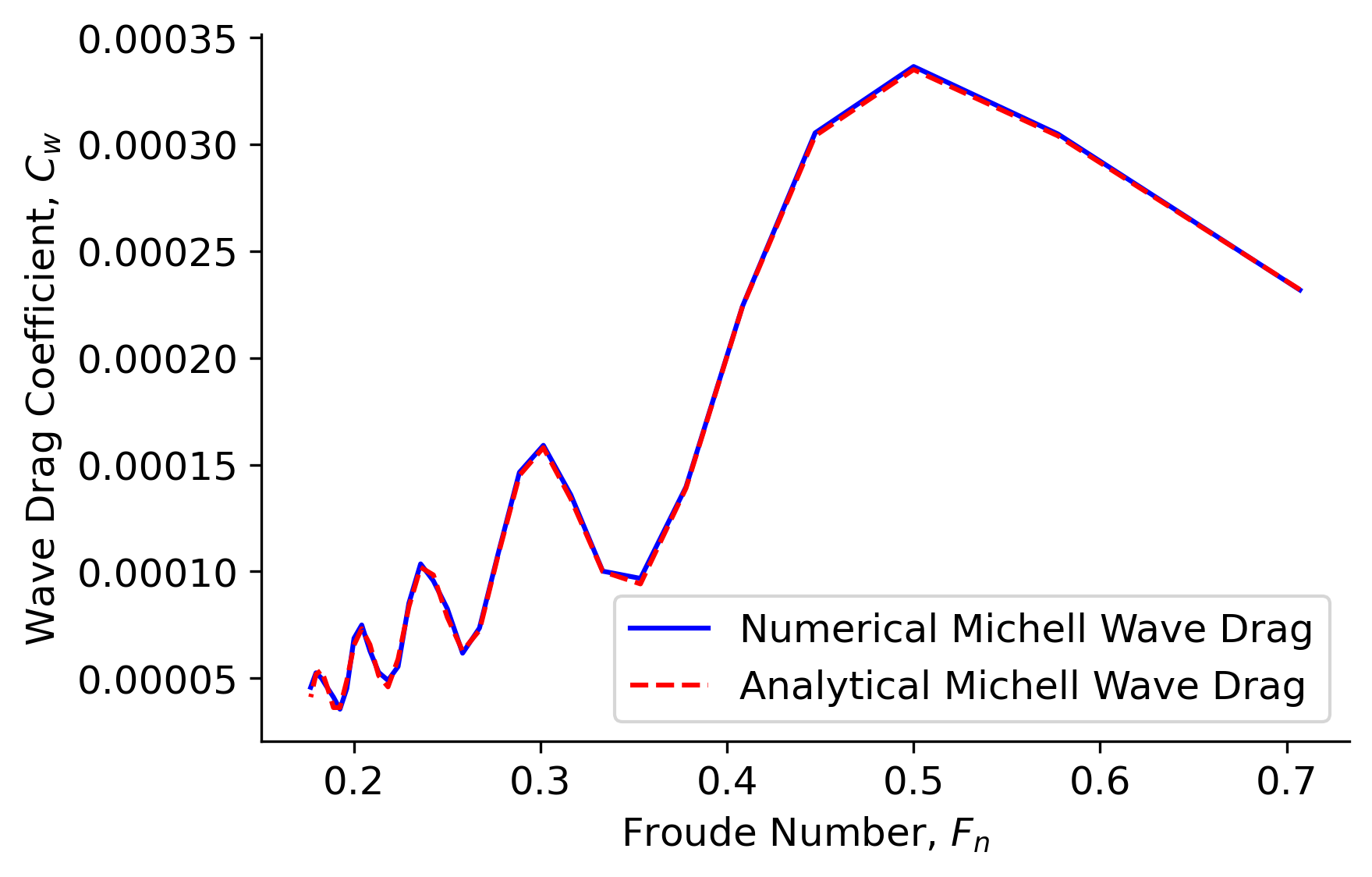}
\caption{The numerical calculation of the Michell integral measured using 301 discrete lengthwise points over 51 discrete waterlines is well resolved to the analytic solution to the Michell integral of the Wigley hull.}
\label{fig:Wigley} 
\end{center}
\end{figure}

\begin{figure}[t]
\begin{center}
\setlength{\unitlength}{0.012500in}%
\includegraphics[width = 3.5in]{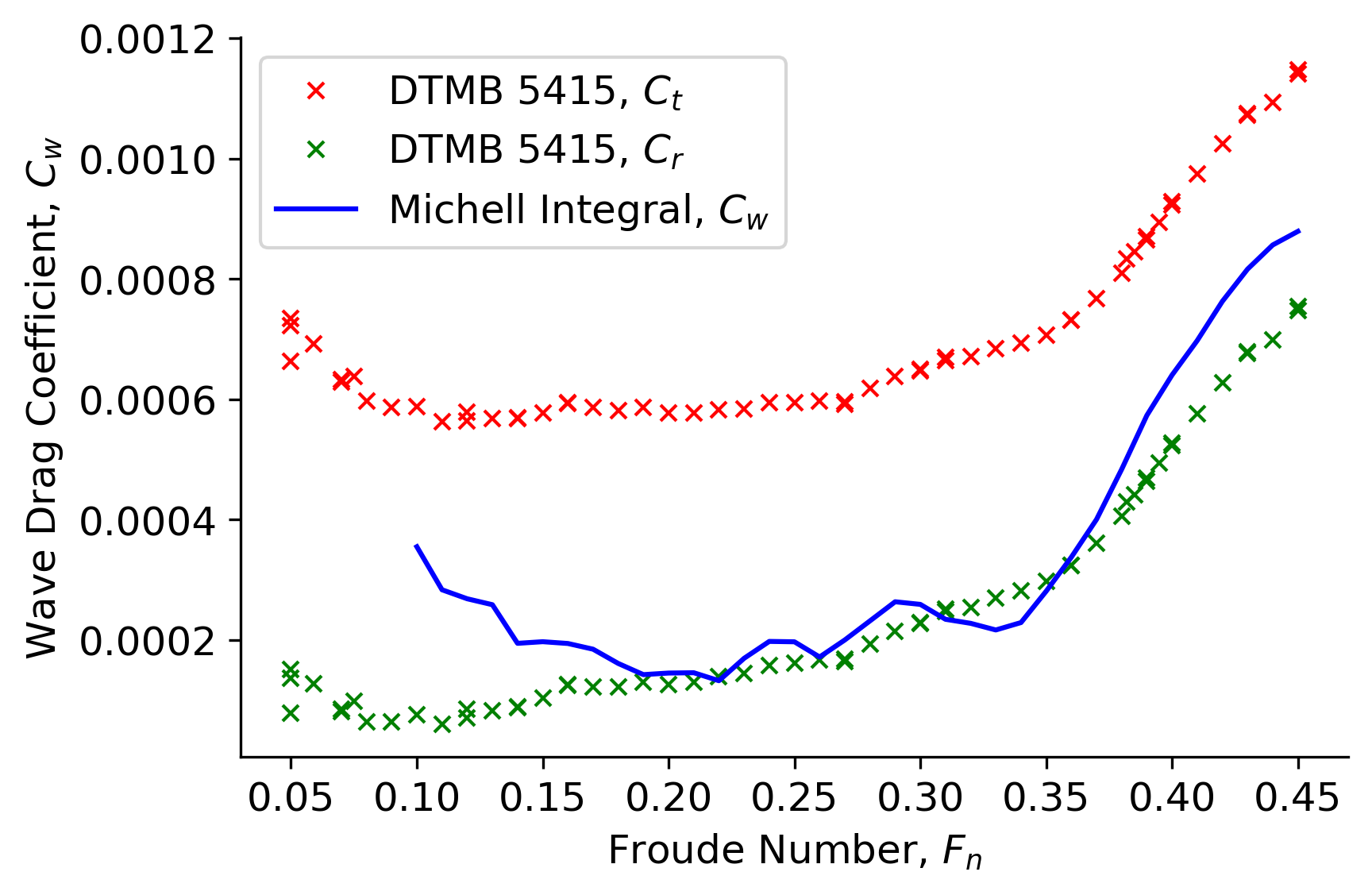}
\caption{The numerical calculation of wave drag coefficient using the Michell integral is well resolved to towing tank measures of the residual resistance coefficient of the DTMB 5414 hull at speeds between $F_n$ = 0.15 AND $F_n$ = 0.45.}
\label{fig:DTMB} 
\end{center}
\end{figure}

Two validation checks were conducted to ensure that the numerical Michell Integral simulation used in the dataset of wave drag coefficients is reasonable. The first test checked the accuracy of the wave drag numerical prediction to the analytical evaluation of wave drag using the Michell Integral. This check was performed using the Wigley hull, a hull with parabolic curvature. Figure~\ref{fig:Wigley} plots the numerical wave drag versus several hull speeds calculated by integrating over 301 discrete points along the length of the hull for 51 waterlines along the displaced volume. Also included in this graph is the analytic solution to the Wigley hull at the same speed conditions\cite{noblesse1983proceedings}. The graph shows that the numerical solution is well resolved to the analytic solution to the Michell Integral. For this reason, the wave drag evaluation for the hull dataset was also computed over a grid of 301 waterline points and 51 waterlines. The second validation performed was to compare the wave drag calculated by the Michell Integral at several speeds to towing tank testing results of a real hull form. Figure~\ref{fig:DTMB} shows that the residual resistance coefficient of the DTMB 5415 hull~\cite{olivieri2001towing} is well resolved to the numerical calculation of the wave drag coefficients using the Michell Integral.

\subsubsection*{Surrogate Model for Resistance Prediction}
A major benefit of creating a dataset is that a surrogate model can be trained to predict the drag of a hull with increased computational speed for a small loss in accuracy. As opposed to constructing a 3D hull from the parameterization and simulating the wave drag, a surrogate model predicts the wave drag coefficients directly from the parameterization vector. The increased speed in drag prediction enables design optimization of a hull on a time scale many orders of magnitude faster than directly simulating the hull. Since the dataset contains hull parameterizations that fall within the full range of feasible hull forms, any feasible hull can have its wave drag coefficients predicted by the surrogate model from its parameterization. The surrogate model used in a regression to predict wave drag coefficients from a hull's parameterization was a residual neural network, chosen for both its speed and the ability to fully differentiate the model.  This means that the derivative of the wave drag coefficient can be taken against any of the terms in the parameterization. 

In the training of the surrogate model, two considerations relating to the distribution of the data were implemented in the training. In the Results Section, Figure~\ref{fig:CW_Dist} shows that the distribution of wave drag coefficients spans several orders of magnitude. Instead of predicting the wave drag coefficient, the surrogate model will predict the $Log_{10}$($C_w$) for the thirty two speed and draft conditions to normalize the final prediction layer. The second consideration implemented in training was to up-sample the instances of hulls that had a wave drag coefficient less than one standard deviation below the mean of samples in the dataset by a factor of four. This up-sampling is intended to increase the prediction accuracy of wave drag prediction for low drag hulls during hull optimization using the trained surrogate model.

\begin{figure}[t]
\begin{center}
\setlength{\unitlength}{0.012500in}%
\includegraphics[width = \columnwidth]{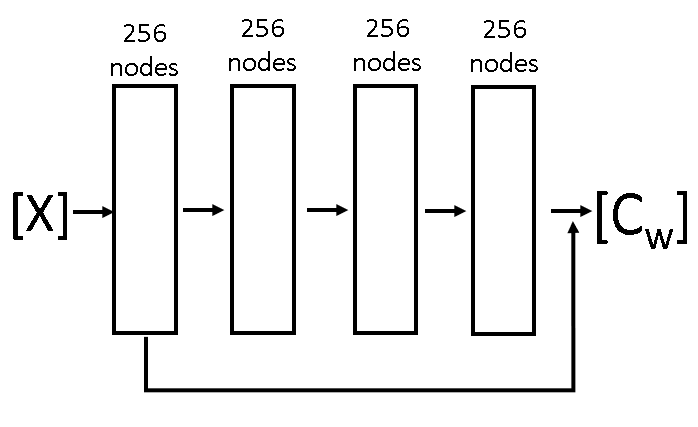}
\caption{The residual neural network trained to predict 32 wave drag coefficients contains 4 hidden layers with 256 nodes in each layer. The input to the ResNet is the hull parameterization and the output is the 32 wave drag coefficients.}
\label{fig:ResNet} 
\end{center}
\end{figure}

After experimenting with different neural network structures, a residual neural network with four hidden layers and 256 nodes in each layer was found to have the greatest prediction accuracy, with an $R^2$ value equal to 0.969. This network structure is seen in Figure~\ref{fig:ResNet}. Immediately prior to the final prediction of the wave drag coefficients, the values in the first hidden layer are summed with the values in the final hidden layer. This assists in boosting the gradients across the network during training to improve the accuracy of the network as a regression model. This residual network (ResNet) predicts the common logarithm of the thirty two wave drag estimates for a given hull in 0.15 seconds, a 480x speed up in for the prediction and is fully differentiable. The ResNet surrogate model is used in a later section of the paper as a tool in hull optimization to minimize drag. 

\section*{RESULTS}
This section details the results of the dataset generation and evaluation. This first subsection provides results for optimizing the parameterization terms to match a set of target hulls. The second subsection provides some statistics and analysis on the dataset of thirty thousand hulls. Finally, the third subsection provides the results of the optimization of the hull parameterization to minimize drag on a hull. 

\subsection*{Hull Matching Validation}
\begin{figure*}[ht]
\begin{center}
\setlength{\unitlength}{0.012500in}%
\includegraphics[width = \linewidth]{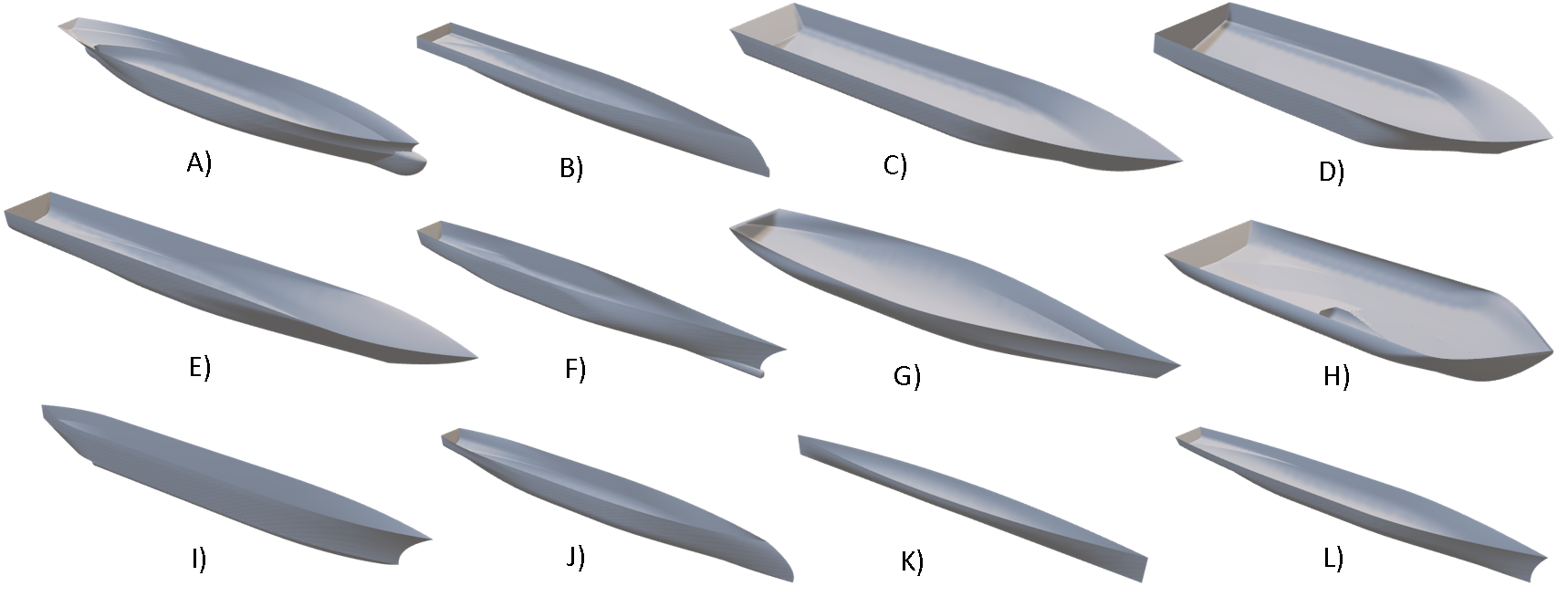}
\caption{The set of twelve parameterized hulls visualize accurate reconstruction of the twelve target hulls.}
\label{fig:ReconHulls} 
\end{center}
\end{figure*}

This subsection provides the results of the parameterized hull matching to the set of twelve target hulls. Figure~\ref{fig:ReconHulls} shows twelve parameterized hulls that were constructed by minimizing the bidirectional mean of the squared Chamfer distance between each parameterized hull and its corresponding target hull from Figure~\ref{fig:TargetHulls}. Table\ref{tab:MatchOpt} lists each of these reconstructed hulls and the normalized square root of the bidirectional mean squared Chamfer distance (RMS of CD). The results of each of these reconstructed hulls vary greatly in scale, so the RMS of CD is listed as a percentage of the LOA of each reconstructed hull. 

\begin{table}[!htbp]
    \centering
    \begin{tabular}{llr}
    \hline
\textbf{Label} & \textbf{Target Hull Name} & \textbf{RMS of CD}\\
\hline
A) & Container Ship~\footnotemark[1]  & 0.343\% \\
B) & \emph{USS Zumwalt}~\footnotemark[2] & 0.252\% \\
C) & Fast Ferry~\footnotemark[3] & 0.277\% \\
D) & Recreational Fishing Boat~\footnotemark[4]  & 0.505\% \\
E) & \emph{USS Freedom}~\footnotemark[5]  & 0.402\% \\
F) & \emph{USS Nimitz}~\footnotemark[6]  & 0.342\% \\
G) & Sailing Yacht~\footnotemark[7]  & 0.397\% \\
H) & Tug Boat~\footnotemark[8]  & 0.559\% \\
I) & \emph{USS Indianapolis}~\footnotemark[9]  & 0.308\% \\
J) & X-Bow Ship~\footnotemark[10]  & 0.390\% \\
K) & Wigley Hull~\cite{noblesse1983proceedings} &  0.0802\% \\
L) & DTMB 5415 Hull~\footnotemark[11]  & 0.401\% \\ 
\hline
    \end{tabular}
    \caption{List of twelve hulls used for reconstruction validation and their corresponding normalized root-mean-square of Chamfer distances. These values are all less than 0.51\% of the hull length, measuring accurate reconstruction. Original hulls are shown in Figure~\ref{fig:TargetHulls}. Reconstructed hulls are shown in Figure~\ref{fig:ReconHulls}}
    \label{tab:MatchOpt}
\end{table}

\footnotetext[1]{https://grabcad.com/library/general-cargo-ship-1}
\footnotetext[2]{https://grabcad.com/library/ddg-1000}
\footnotetext[3]{https://grabcad.com/library/fast-passenger-monohull-ferry-mh35-1}
\footnotetext[4]{https://grabcad.com/library/10-meters-fishing-boat-1}
\footnotetext[5]{https://grabcad.com/library/littoral-combat-ship-1}
\footnotetext[6]{https://grabcad.com/library/hull-of-carrier-of-nimitz-class-1}
\footnotetext[7]{https://grabcad.com/library/36-meter-sailing-yacht-1}
\footnotetext[8]{https://grabcad.com/library/renko-dangar-marine-steel-boat-project-1}
\footnotetext[9]{https://grabcad.com/library/uss-indianapolis-ca-35-1}
\footnotetext[10]{https://grabcad.com/library/x-bow-hull-1}
\footnotetext[11]{http://www.simman2008.dk/5415/5415\_geometry.htm}

\subsection*{Dataset of Parameterized Hulls}
\begin{figure}[t]
\begin{center}
\setlength{\unitlength}{0.012500in}%
\includegraphics[width = 3.5in]{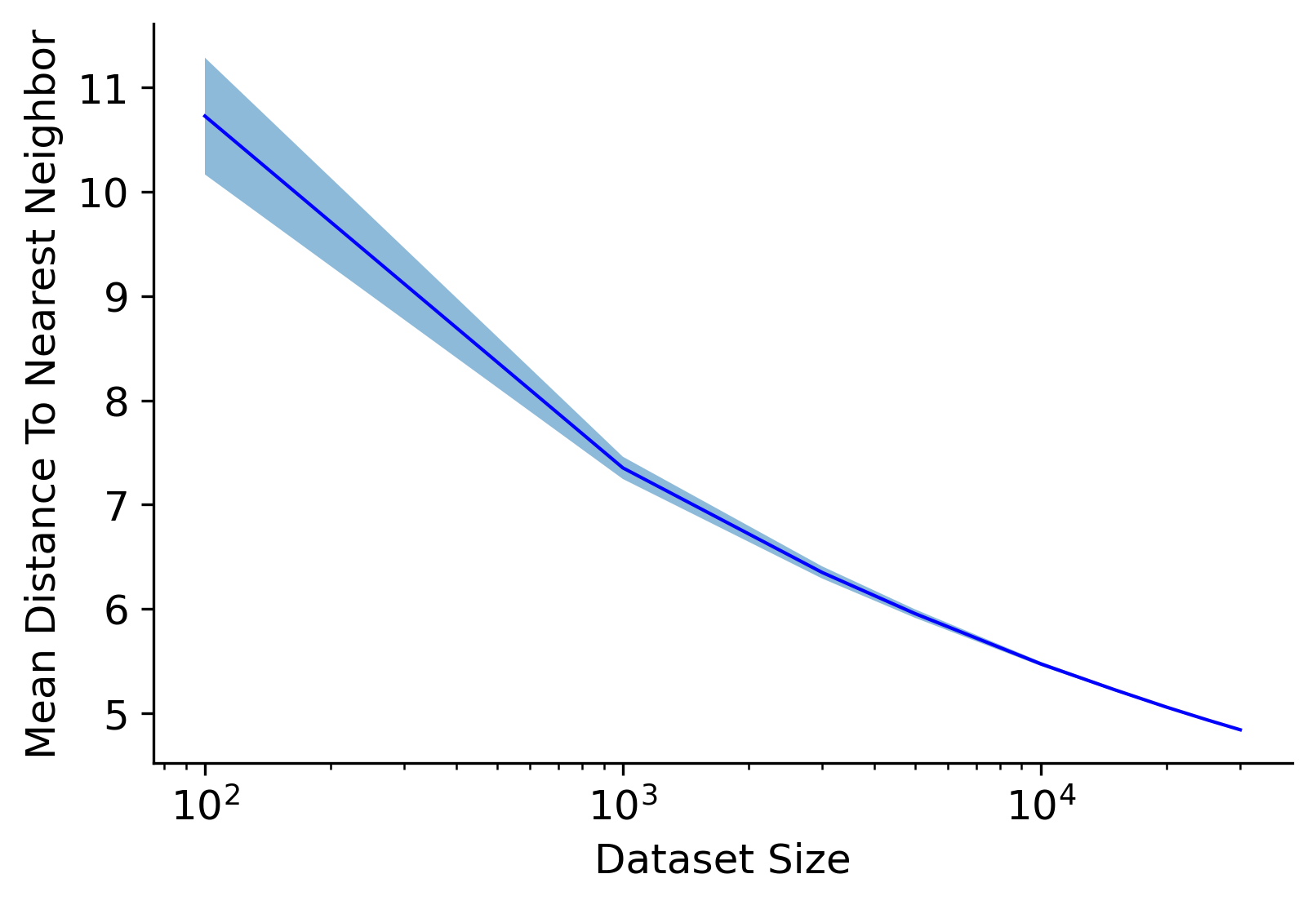}
\caption{Mean Euclidean distance to nearest neighbor with increasing sample size. The shaded region shows 2 standard deviations from the mean collapses with increasing sample size.}
\label{fig:KNN_Dist} 
\end{center}
\end{figure}

After generating thirty thousand hulls and computing the thirty two wave drag coefficients, some analysis was performed on the dataset to better understand the parameterization distribution across the feasible region in the design space and extreme minimum values associated with measurements of drag. One measure of the distribution of the spread of hull samples is the average Euclidean distance each hull parameterization vector is to its nearest neighbor. Figure~\ref{fig:KNN_Dist} shows this measure with increasing sample size in the dataset. Further analysis of the spread of samples in the dataset is provided in the Discussion Section. 

\begin{figure}[t]
\begin{center}
\setlength{\unitlength}{0.012500in}%
\includegraphics[width = \columnwidth]{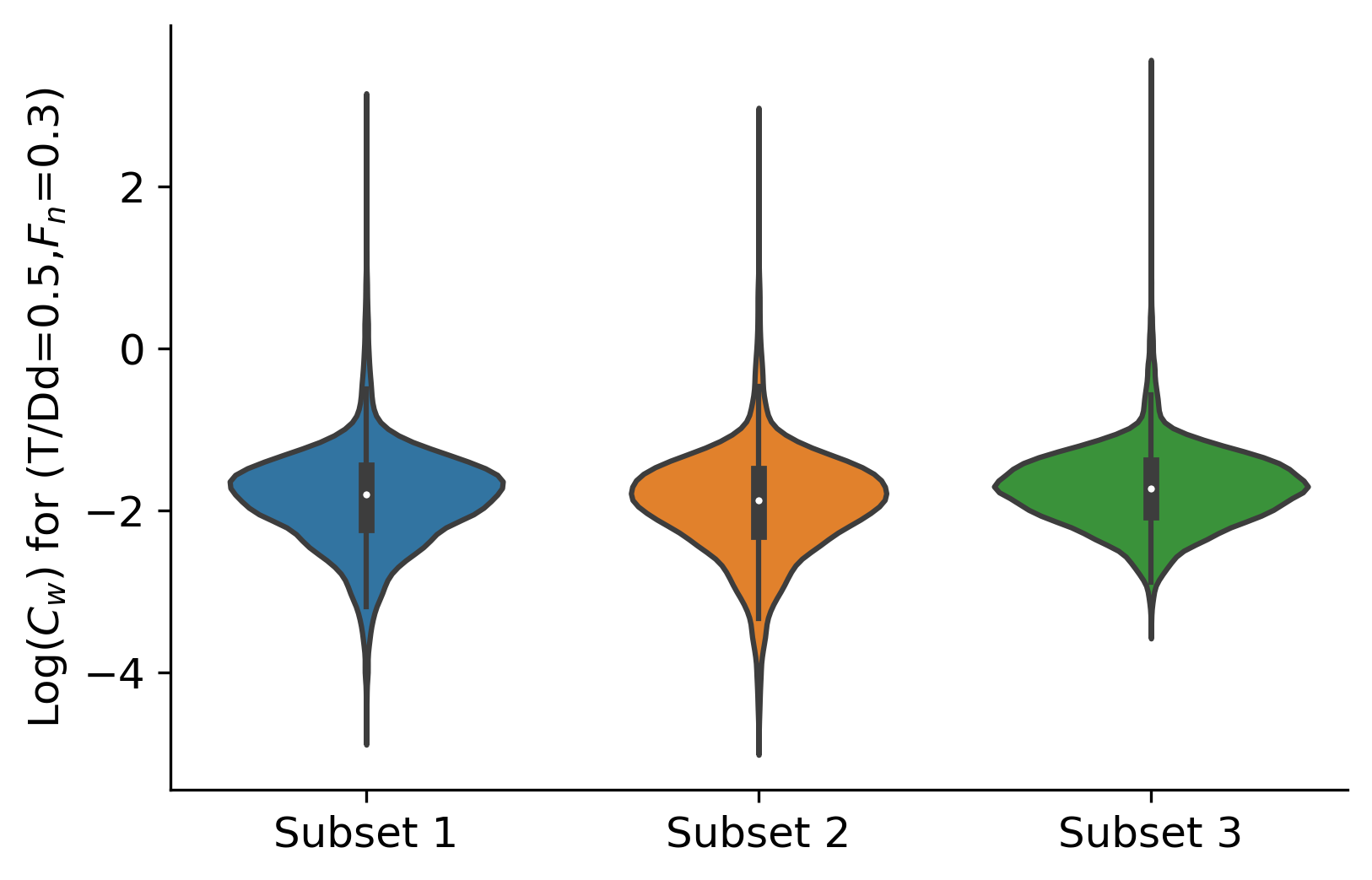}
\caption{Violin plots of wave drag coefficients of dataset hulls for a draft of 1/2 the depth of the hull and a Froude number = 0.3 ($(F_N = 0.3, T/Dd = 0.5)$). Results are separated into the dataset subsets}
\label{fig:CW_Dist} 
\end{center}
\end{figure}

Another measure of the dataset of hulls is to measure the spread of the wave drag coefficients of the samples. Figure~\ref{fig:CW_Dist} shows the distribution of the wave drag coefficients across the three subsets in the dataset. It is important to note that the scale of the Y-axis of this chart is on a logarithmic scale as the distribution of wave drag coefficients spans multiple orders of magnitude. Please see the Discussion Section for an analysis of this result. 

\begin{figure}[t]
\begin{center}
\setlength{\unitlength}{0.012500in}%
\includegraphics[width = \columnwidth]{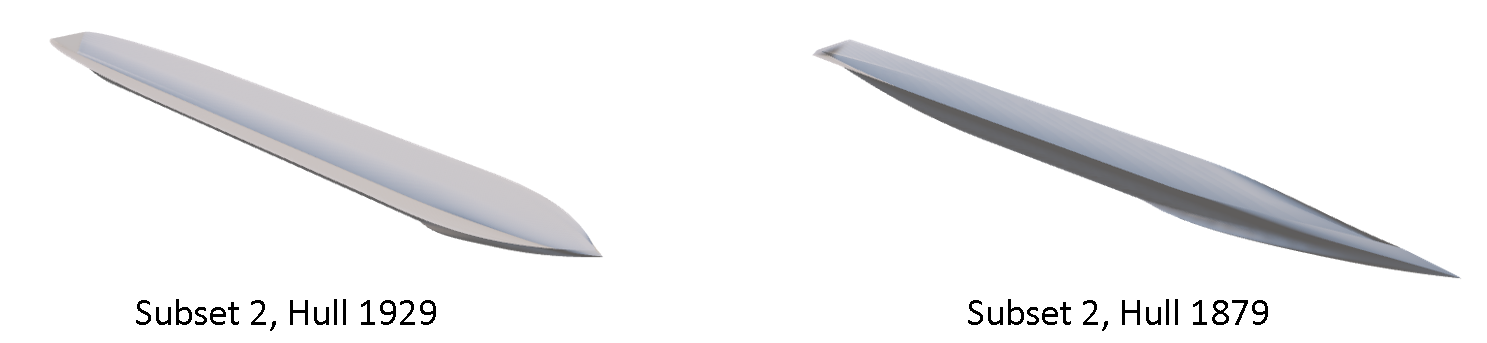}
\caption{Hulls from the dataset with the minimum wave drag coefficients. The left hull has the lowest wave drag coefficient at the $(F_N = 0.3, T/Dd = 0.5)$ condtions. The right hull has the lowest aggregated wave drag coefficients across all speed conditions at $T/Dd = 0.5$.}
\label{fig:LowCW hulls} 
\end{center}
\end{figure}

\begin{figure}[t]
\begin{center}
\setlength{\unitlength}{0.012500in}%
\includegraphics[width = \columnwidth]{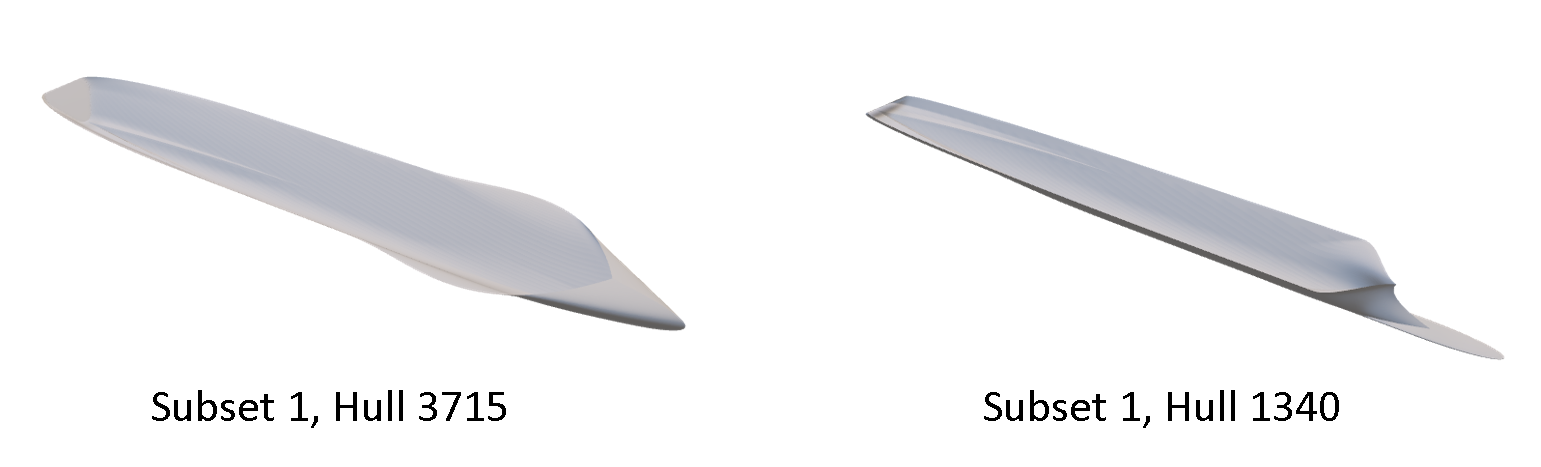}
\caption{Hulls from the dataset with the lowest total drag for $U = 25 knots$ when scaled to different volumetric displacements. The left hull has the lowest total drag when scaled to displacement volumes of 5000, 10000, and 100000 cubic meters. The right hull has the lowest total drag when scaled to displacement volumes of 500 and 1000 cubic meters.}
\label{fig:Low Rt Hulls} 
\end{center}
\end{figure}

Minimum values of different measures of drag in the dataset were also collected. One measure of the dataset of hulls was to find the hulls with the minimum total drag when scaled to different volumetric displacements. Table~\ref{tab:MinRT} showcases that two different hulls standout among the dataset as having the lowest total drag. These two hulls can be seen in Figure~\ref{fig:Low Rt Hulls}. Hull 1-3715 has the lowest total drag for displacements ranging from 5000 cubic meters to 100000 cubic meters, while Hull 1-1340 has the lowest drag for displacements of 500 cubic meters and 1000 cubic meters. As scale decreases, viscous forces increase suggesting why different hulls at different scaled displacements have minimum drag. In addition to looking at total drag, hulls from the dataset with the lowest coefficients were also found. The first case was the hull with the lowest drag at a speed-draft condition of $(F_N = 0.3, T/Dd = 0.5)$. This hull had a wave drag coefficient equal to $1.53*10^{-5}$. The second measure of minimum wave drag was to aggregate the wave drag coefficients across all tested speeds at a $T/Dd = 0.5$ condition. Both hulls are shown in Figure~\ref{fig:LowCW hulls}. Analysis of the dataset as it pertains to drag is provided in the Discussion Section.

\begin{table}[!htbp]
    \centering
    \begin{tabular}{lccc}
    \hline
\textbf{DISP. VOLUME} & \textbf{$R_T$} & \textbf{HULL} & \textbf{LOA}\\
\hline
$500m^3$ & $67.04kN$ & Set 1, \#1340 & $109.38m$\\
$1000m^3$ & $96.97kN$ & Set 1, \#1340 & $137.81m$\\
$5000m^3$ & $284.0kN$ & Set 1, \#3715 & $225.09m$\\
$10000m^3$ & $338.2kN$ & Set 1, \#3715 & $283.59m$\\
$100000m^3$ & $1046kN$ & Set 1, \#3715 & $611.0m$\\
\hline
    \end{tabular}
    \caption{Tabulated data of the dataset hulls with minimum total drag when scaled to different volumetric displacements. Skin friction and wave drag scale differently, leading to different minimal drag hulls at different scales.}
    \label{tab:MinRT}
\end{table}

\subsection*{Hull Form Optimization Via ResNet Surrogate for Wave Drag Coefficient}
\begin{figure}[t]
\begin{center}
\setlength{\unitlength}{0.012500in}%
\includegraphics[width = \columnwidth]{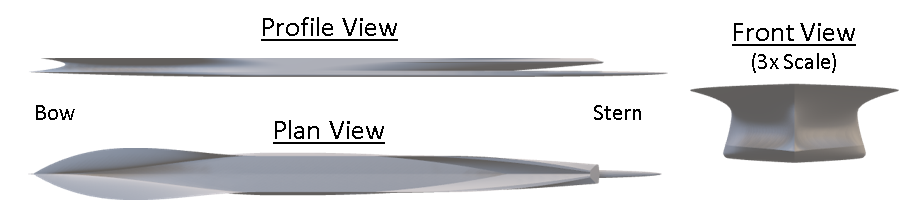}
\caption{Optimized hull that displaces 100000 cubic meters has 25\% reduction in drag compared to the hull in the dataset with the minimum drag in the same conditions}
\label{fig:minRT hull} 
\end{center}
\end{figure}

Optimization of the hull parameters saw significant reductions in total drag for three test cases. The first optimization of the parameters was constrained so that the hull had a volumetric displacement of 100000 cubic meters and a speed of 25 knots. The optimized hull had a total drag of 785kN, which is a 25\% Reduction in total drag compared to the hull in the dataset with the lowest drag. Results were computed with NSGA2 to minimize both $R_t$ and the interpolated $C_w$ for the speed/draft condition. The hulls in the final population were then constructed and total drag was measured with the Michell Integral for the calculation of wave drag. The hull shown in Figure~\ref{fig:minRT hull} shows this optimized hull, which had a total length of 685.24 meters. It is important to note that this hull is not the hull with the minimum drag predicted by the ResNet, but it did belong to the final population of optimized hulls. 

\begin{figure*}[ht]
\begin{center}
\setlength{\unitlength}{0.012500in}%
\includegraphics[width = \linewidth]{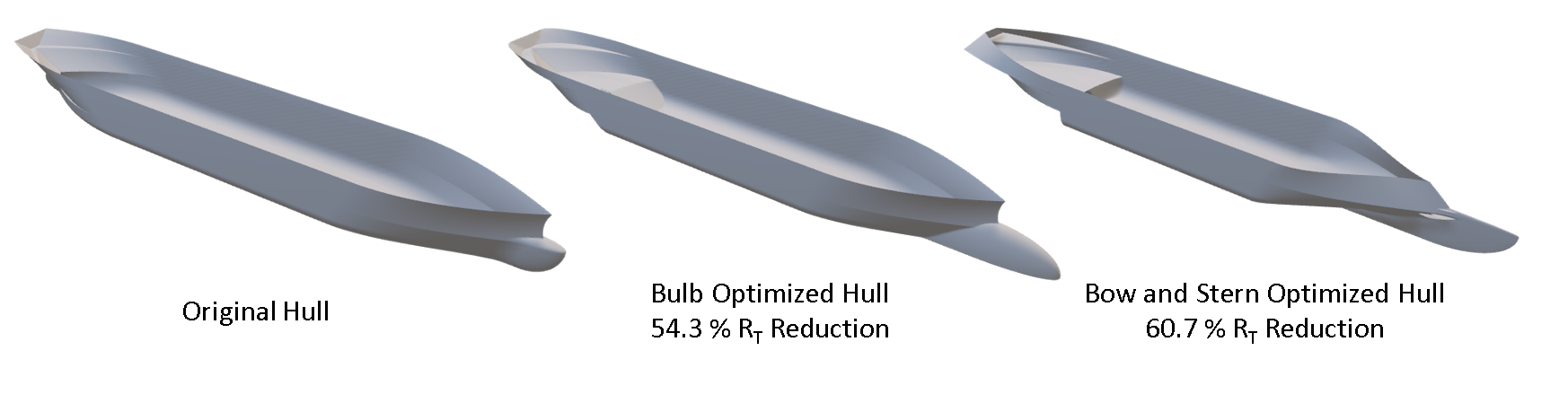}
\caption{Optimization of the modified container ship created constrained to only optimizing the bulbs on the hull lead to a 54.3\% reduction in total drag. Optimization parameters relating to the shape of the bow taper, stern taper, and bulbs lead to a 60.7\% reduction in total drag.}
\label{fig:OpHulls} 
\end{center}
\end{figure*}

The second and third optimizations were modeled after problems found in the literature. The goal was to optimize a hull while maintaining the aspect of the geometry of an initial hull form. This initial hull the container ship hull from Figure~\ref{fig:ReconHulls} was re-proportioned to have an LOA of 200 meters and a bow and stern taper over the forward-most and aft-most 30\% of the hull. The total drag on this hull with an operating speed of 25 knots and a draft of 12.5 meters is $2.694*10^7N$. With the same optimization process described above, two optimizations were performed. The first optimization only manipulated the parameters related to the bulb geometries. This optimization reduced the total drag on the hull by 54.3\% in the same speed and draft condition. A second optimization manipulated the parameters associated with the bow, stern, and bulbs while maintaining the body of the midship. This optimization reduced the total drag of the hull by 60.7\% for the same speed and draft condition. Figure~\ref{fig:OpHulls} shows the original container ship hull, the same hull with optimized bulbs, and the optimized hull with the same parallel midbody as the initial hull. Further analysis of the hull optimization with the ResNet is provided in the Discussion Section. 

\section*{DISCUSSION}
This section analyzes the findings from the Results Section. The first subsection provides insight into the reconstruction of target hulls as parameterized hulls. The second subsection analyzes the dataset of parameterized hulls and the distribution of information provided in the set. The third subsection provides an analysis on the training of the ResNet to predict the wave drag coefficients of a hull at multiple operating conditions. In conjunction with the third subsection, the fourth subsection analyzes the ResNet's ability as a regression model to be used with an optimization algorithm to minimize drag. 

\subsection*{Parameterized Reconstruction of Target Hulls}
Based on the visual and computational results of this study, the parameterization scheme proposed in this paper can accurately reconstruct a large variety of classical hull forms. The creation of this parameterization was specifically aimed at closing two specific gaps found in the literature:
\begin{enumerate}
    \item A comprehensive design representation of ship hulls that encompasses a large diversity of geometric features seen on hulls
    \item A design representation for ship hulls that allows for human manipulation and computational design methods
\end{enumerate}
The parametric scheme was designed to allow human input in terms of concrete geometric features seen in ship hulls. Additionally, this scheme allows for computer generated inputs so that a ship hull can be computationally designed in the same data frame as the human defined design. Due to this, future machine learning models leveraging this parametric scheme for ship hull design can learn from both human and computer generated inputs. These results will serve as a model for future benchmark studies for comprehensive parametric reconstruction of ship hulls. 

The parametric scheme is certainly not a panacea to parametric ship hull design, as there are a few features seen in the target hulls that this parameterization is able to reasonably reconstruct. Most notably, the target hull for a fishing boat (image D in Figures~\ref{fig:TargetHulls} and~\ref{fig:ReconHulls}) has a stepped hull and longitudinal strakes for assisting in the hull's hydrodynamics. The surface of the  reconstructed parameterized hull of the fishing boat is smooth as the parameterization cannot create these features. Another example, the target hull for the \emph{USS Nimitz}(image F in Figures~\ref{fig:TargetHulls} and~\ref{fig:ReconHulls}) has a flared, non-pointed, bow to support structure for the aircraft runway on the ships's deck. This geometric feature is impossible with the proposed parameterization. The optimized reconstructed parameterized hull of the \emph{USS Nimitz} minimized the Chamfer distance over a majority of the hull surface but had a pointed bow form as defined with the parameterization. A third notable feature is that the parameterization had difficulty reconstructing the bulb in the DTMB 5415 (image L in Figures~\ref{fig:TargetHulls} and~\ref{fig:ReconHulls}) target hull has the bulb protruded below the baseline of the hull. This positioning of the bulb is not possible with the parameterization. It is important to note that the reconstructed parameterized DTMB hull does not contain a bulb at all.  

Overall, this parameterization can comprehensively design diverse hulls for human and machine learning design processes. The 45 parameters defined in this parameterization provide sufficient information to reconstruct a large array of hull geometries, yet does not over-complicate the design space in a way that inhibits its usefulness in a computational model due to the curse of dimensionality. This parameterization is shown to design hull geometries seen across the full spectrum of traditional hull designs and scales, from small recreational watercraft to large naval hulls. One note to add is that the geometry of a hull is defined by 44 parameters and the final parameter is the hull's length overall. This way, in theory, the geometry of the recreational fishing boat (Hull D) could exist as any length by only changing the length overall parameter. Additionally, the parameterization allows for future machine learning work performed on the dataset to focus on tools and methods for performance prediction and design generation without the need to create models for representing the design of a hull given only its mesh. Future work in the hull representation will look at reconstructing minute features that are specific to certain types of ships. 

\subsection*{Data Set Generation}
The dataset of thirty thousand ship hulls was generated to cover the full range of feasible samples for the hull parameterization. In order to ensure that the total feasible design space of ship hulls is encompassed in the dataset, the Euclidean distance to the nearest neighbor of each parameterized hull was measured with increasing dataset size. Figure~\ref{fig:KNN_Dist} shows that as the number of samples approaches thirty thousand the slope of this curve and the standard deviation significantly decays, suggesting that the entire design space of hulls is well sampled with thirty thousand samples. More samples will certainly yield better results; however the current sampling achieved a point of diminishing returns, where any further reduction in the mean distance will require a significant increase in the number of samples.

While the dataset covers the design space of feasible hulls well, this produced a large range of wave drag coefficients in the performance space, spanning several orders of magnitude. Figure~\ref{fig:CW_Dist} shows the distribution of wave drag coefficients across the three subsets of hulls in the dataset, showing that they all have similar means. It is important to note that the distribution of wave drag coefficients for subset 3 is narrower and skewed higher than that of subsets 1 and 2. From a design perspective, this is a reasonable outcome. The hulls in subset 3 are intended to represent large hulls, especially for shipping. The design of larger hulls is dependent on both the hydrodynamics the ability to carry large amounts of cargo. Due to this required balance, it makes sense that a set of designs skewed to represent cargo ships would have higher drag coefficients as there are additional design criteria for these hull forms. In addition, subset 2 was skewed to represent hull forms seen on smaller ships. So, it also makes sense that these hulls might have lower drag coefficients as the design of smaller craft is dominated by speed, whether this is for competition, leisure, or other purposes. 

The design space coverage and large distribution of wave drag coefficients led to the successful training of the ResNet to learn how the geometric features of a hull defined by the parameterization can accurately predict the drag of a hull. The next subsection discusses the implications of the surrogate model training for design evaluation and optimization. 

\subsubsection*{Hull Form Optimization for Drag via ResNet}
Using the ResNet to quickly predict drag within an optimization of a hull led to the generation of hulls with significantly reduced drag. Minimizing total drag with few constraints produces a hull with a 25\% reduction in drag compared to the hull in the dataset with the minimal drag with a displaced volume of 100000 cubic meters. Unfortunately, the optimized hull is too long to be a real ship. For reference, one of the target hulls provided in the paper, the \emph{USS Nimitz}, has a displacement of approximately 1000000 cubic meters, but is only 332 meters in length. Due to constraints on the global infrastructure of ports, very few ships exceed 350 meters in length. This optimized hull is more than double the length of the \emph{USS Nimitz}, yet displaces approximately the same volume, suggesting that there are harder constraints on the design of real ships than only drag. This same argument was detailed in the analysis of the distributions of data across the three subsets in the previous section. An additional consideration in the design of a ship with the optimized hull is the increased structural need of supporting bending moments on this hull, further limiting its mission capabilities. Neglecting that the optimization produced a practically infeasible hull, the optimization with the ResNet had two important findings:
\begin{enumerate}
    \item The ResNet was able to learn how geometric features of hulls affect drag
    \item Hull optimization with the ResNet was able to find geometric features in combination and was able to reduce drag more than any of the hulls existing in the training data.
\end{enumerate}
Additionally, constrained optimization of an initial hull form showed that wave drag coefficient predictions with the ResNet were able to consider the individual geometric features of the hull parameterization and optimize parameters to produce ship hulls with significantly reduced drag compared to the initial hull form. This is true for optimizing local features such as bulbs, or global features such as the entire bow and stern of the hull. It is important to note that the optimization of a bulbous bow is a delicate balancing act. The bulb creates destructive interference in a hull's wave, reducing the wave drag; however, this is at the cost of increasing skin friction. The optimization with the ResNet balanced this trade-off. A reduction in 5-10\% of total drag is no small feat for a human hull designer to accomplish.  Computational tools reducing the drag of a hull upwards of 60\% will certainly produce greatly improved outcomes in the cost of shipping and fossil fuel emissions. The current limit in this potential is in creating ship hulls that are practically feasible for real-world use. 

\section*{CONCLUSION}
This paper describes the creation of a ship hull dataset for computational and data-driven design. The dataset was generated using a novel parameterization method that comprehensively covers the vast design space of ship hulls, including traditional geometries. This parameterization method allows for both human and computer-generated designs to exist within the same data frame. It accurately reconstructs 12 distinct ship hulls with a diversity of geometric features, with a normalized root-mean-square of Chamfer distance of less than 0.51\% of the target hull's length. The resulting dataset contains 30,000 ship hulls covering the full design space of feasible hull geometries, with some bias towards realistic hull features. This dataset is over 42 times larger than any other publicly available ship hull dataset and characterizes more geometric features. A surrogate model based on ResNet architecture is trained on this dataset that accurately predicts the wave drag coefficients for 32 speed/draft conditions, with an R-squared value of 0.969. A case study of surrogate-based optimization to minimize total drag of a hull in constrained conditions demonstrates the model's ability to predict the influence of individual geometric features and the influence of high-quality geometric features in combination, resulting in reductions of up to 60\% in total hull drag.

Future work will involve characterizing the performance of ship hulls using additional performance metrics, including geometric, hydrostatic, and hydrodynamic measures. In addition, a future study into surrogate modeling with multi fidelity simulation will explore surrogate prediction accuracy and computational effort to produce datasets for early stage data driven design. Another consideration for future work is that many hulls in the current dataset have high drag. This indicates that the random sampling of parameters may not be an effective approach for generating high-performing hulls. Future work will aim to generate a larger dataset that considers hull performance in addition to geometric feasibility, enabling the training of surrogate models with greater accuracy in the regions of the total design space containing high-performing hulls similar to existing real hull forms.

\section*{ACKNOWLEDGEMENTS}
 We thank the National Defense Science and Engineering Graduate Fellowship program for supporting Noah Bagazinski's studies while working on this project. Additionally, we thank MIT Supercloud~\cite{supercloud} for providing the computational resources needed for the work performed in this paper. The dataset, code, and documentation for this work is at \url{https://github.com/noahbagz/ShipD}.

\bibliographystyle{asmems4}

%
% ...

\bibliography{asme2e}

\begin{thebibliography}{10}

\bibitem{chen2021padgan}
Chen, W., and Ahmed, F., 2021.
\newblock ``Padgan: Learning to generate high-quality novel designs''.
\newblock {\em Journal of Mechanical Design, \textbf{ 143}}(3).

\bibitem{mirhoseini2021graph}
Mirhoseini, A., Goldie, A., Yazgan, M., Jiang, J.~W., Songhori, E., Wang, S.,
  Lee, Y.-J., Johnson, E., Pathak, O., Nazi, A., et~al., 2021.
\newblock ``A graph placement methodology for fast chip design''.
\newblock {\em Nature, \textbf{ 594}}(7862), pp.~207--212.

\bibitem{evans1959basic}
Evans, J.~H., 1959.
\newblock ``Basic design concepts''.
\newblock {\em Journal of the American Society for Naval Engineers, \textbf{
  71}}(4), pp.~671--678.

\bibitem{lin2017feature}
Lin, C.-K., and Shaw, H.-J., 2017.
\newblock ``Feature-based estimation of preliminary costs in shipbuilding''.
\newblock {\em Ocean Engineering, \textbf{ 144}}, pp.~305--319.

\bibitem{regenwetter2022biked}
Regenwetter, L., Curry, B., and Ahmed, F., 2022.
\newblock ``Biked: A dataset for computational bicycle design with machine
  learning benchmarks''.
\newblock {\em Journal of Mechanical Design, \textbf{ 144}}(3).

\bibitem{regenwetter2022framed}
Regenwetter, L., Weaver, C., and Ahmed, F., 2023.
\newblock ``Framed: An automl approach for structural performance prediction of
  bicycle frames''.
\newblock {\em Computer-Aided Design, \textbf{ 156}}, p.~103446.

\bibitem{heyrani2022links}
Heyrani~Nobari, A., Srivastava, A., Gutfreund, D., and Ahmed, F., 2022.
\newblock ``Links: A dataset of a hundred million planar linkage mechanisms for
  data-driven kinematic design''.
\newblock In International Design Engineering Technical Conferences and
  Computers and Information in Engineering Conference, Vol.~86229, American
  Society of Mechanical Engineers, p.~V03AT03A013.

\bibitem{chan2021metaset}
Chan, Y.-C., Ahmed, F., Wang, L., and Chen, W., 2021.
\newblock ``Metaset: Exploring shape and property spaces for data-driven
  metamaterials design''.
\newblock {\em Journal of Mechanical Design, \textbf{ 143}}(3).

\bibitem{lee2023tMetaset}
Lee, D., Chan, Y.-C., Chen, W., Wang, L., van Beek, A., and Chen, W., 2023.
\newblock ``t-metaset: Task-aware acquisition of metamaterial datasets through
  diversity-based active learning''.
\newblock {\em Journal of Mechanical Design, \textbf{ 145}}(3), p.~031704.

\bibitem{read2009drag}
Read, D., 2009.
\newblock {\em A drag estimate for concept-stage ship design optimization}.
\newblock The University of Maine.

\bibitem{song2023attention}
Song, B., Miller, S., and Ahmed, F., 2023.
\newblock ``Attention-enhanced multimodal learning for conceptual design
  evaluations''.
\newblock {\em Journal of Mechanical Design}, pp.~1--38.

\bibitem{maze2022topodiff}
Maz{\'e}, F., and Ahmed, F., 2022.
\newblock ``Topodiff: A performance and constraint-guided diffusion model for
  topology optimization''.
\newblock {\em arXiv preprint arXiv:2208.09591}.

\bibitem{brown2003multiple}
Brown, A., and Salcedo, J., 2003.
\newblock ``Multiple-objective optimization in naval ship design''.
\newblock {\em Naval Engineers Journal, \textbf{ 115}}(4), pp.~49--62.

\bibitem{HullOpt}
Feng, Y., el~Moctar, O., and Schellin, T., 2022.
\newblock ``Parametric hull form optimization of containerships for minimum
  resistance in calm water and in waves''.
\newblock {\em Journal of Marine Science and Applications}, January.

\bibitem{khan2022shape}
Khan, S., Kaklis, P., Serani, A., Diez, M., and Kostas, K., 2022.
\newblock ``Shape-supervised dimension reduction: Extracting geometry and
  physics associated features with geometric moments''.
\newblock {\em Computer-Aided Design, \textbf{ 150}}, p.~103327.

\bibitem{khan2022geometric}
Khan, S., Kaklis, P., Serani, A., and Diez, M., 2022.
\newblock ``Geometric moment-dependent global sensitivity analysis without
  simulation data: application to ship hull form optimisation''.
\newblock {\em Computer-Aided Design, \textbf{ 151}}, p.~103339.

\bibitem{zhang2018parametric}
Zhang, Y., Kim, D.-J., and Bahatmaka, A., 2018.
\newblock ``Parametric method using grasshopper for bulbous bow generation''.
\newblock In 2018 International Conference on Computing, Electronics \&
  Communications Engineering (iCCECE), IEEE, pp.~307--310.

\bibitem{chrismianto2014parametric}
Chrismianto, D., and Kim, D.-J., 2014.
\newblock ``Parametric bulbous bow design using the cubic bezier curve and
  curve-plane intersection method for the minimization of ship resistance in
  cfd''.
\newblock {\em Journal of Marine Science and Technology, \textbf{ 19}},
  pp.~479--492.

\bibitem{lu2016hydrodynamic}
Lu, Y., Chang, X., and Hu, A.-k., 2016.
\newblock ``A hydrodynamic optimization design methodology for a ship bulbous
  bow under multiple operating conditions''.
\newblock {\em Engineering Applications of Computational Fluid Mechanics,
  \textbf{ 10}}(1), pp.~330--345.

\bibitem{PSOShip_Opt}
Knight, J.~T., Zahradka, F.~T., Singer, D.~J., and Collette, M.~D., 2014.
\newblock ``{Multiobjective Particle Swarm Optimization of a Planing Craft with
  Uncertainty}''.
\newblock {\em Journal of Ship Production and Design, \textbf{ 30}}(04), 11,
  pp.~194--200.

\bibitem{PSO_Multi_Opt}
Knight, J.~T., Singer, D.~J., and Collette, M.~D., 2015.
\newblock ``{Testing of a spreading mechanism to promote diversity in
  multi-objective particle swarm optimization}''.
\newblock {\em Optimization and Engineering, \textbf{ 16}}, June, pp.~279--302.

\bibitem{wang2022shipEncoding}
Wang, Y., Joseph, J., Aniruddhan~Unni, T., Yamakawa, S., Barati~Farimani, A.,
  and Shimada, K., 2022.
\newblock ``Three-dimensional ship hull encoding and optimization via deep
  neural networks''.
\newblock {\em Journal of Mechanical Design, \textbf{ 144}}(10), p.~101701.

\bibitem{ao2021artificial}
Ao, Y., Li, Y., Gong, J., and Li, S., 2021.
\newblock ``An artificial intelligence-aided design (aiad) of ship hull
  structures''.
\newblock {\em Journal of Ocean Engineering and Science}.

\bibitem{ao2022artificial}
Ao, Y., Li, Y., Gong, J., and Li, S., 2022.
\newblock ``Artificial intelligence design for ship structures: A variant
  multiple-input neural network-based ship resistance prediction''.
\newblock {\em Journal of Mechanical Design, \textbf{ 144}}(9), p.~091707.

\bibitem{peri2001design}
Peri, D., Rossetti, M., and Campana, E.~F., 2001.
\newblock ``Design optimization of ship hulls via cfd techniques''.
\newblock {\em Journal of ship research, \textbf{ 45}}(02), pp.~140--149.

\bibitem{demo2021hull}
Demo, N., Tezzele, M., Mola, A., and Rozza, G., 2021.
\newblock ``Hull shape design optimization with parameter space and model
  reductions, and self-learning mesh morphing''.
\newblock {\em Journal of Marine Science and Engineering, \textbf{ 9}}(2),
  p.~185.

\bibitem{li2022machine}
Li, J., Du, X., and Martins, J.~R., 2022.
\newblock ``Machine learning in aerodynamic shape optimization''.
\newblock {\em Progress in Aerospace Sciences, \textbf{ 134}}, p.~100849.

\bibitem{marlantes2021Modeling}
Marlantes, K., and Maki, K., 2021.
\newblock ``Modeling vertical planing boat motions using a neural-corrector
  method''.

\bibitem{AppNavArch}
Zubaly, R., 1996.
\newblock {\em Applied Naval Architecture}.
\newblock Cornell Maritime Press.

\bibitem{noblesse2013neumann}
Noblesse, F., Huang, F., and Yang, C., 2013.
\newblock ``The neumann--michell theory of ship waves''.
\newblock {\em Journal of Engineering Mathematics, \textbf{ 79}}(1),
  pp.~51--71.

\bibitem{noblesse1983proceedings}
Noblesse, F., and McCarthy, J., 1983.
\newblock Proceedings of the dtnsrdc (david w. taylor naval ship research and
  development center) workshop on ship wave-resistance computations (2nd).
\newblock Tech. rep., David W Taylor Naval Ship Research and Development
  Center, November.

\bibitem{huang2013numerical}
Huang, F., Yang, C., and Noblesse, F., 2013.
\newblock ``Numerical implementation and validation of the neumann--michell
  theory of ship waves''.
\newblock {\em European Journal of Mechanics-B/Fluids, \textbf{ 42}},
  pp.~47--68.

\bibitem{newman2018marine}
Newman, J.~N., 2018.
\newblock {\em Marine hydrodynamics}.
\newblock The MIT press.

\bibitem{yang2013practical}
Yang, C., Huang, F., and Noblesse, F., 2013.
\newblock ``Practical evaluation of the drag of a ship for design and
  optimization''.
\newblock {\em Journal of Hydrodynamics, \textbf{ 25}}(5), pp.~645--654.

\bibitem{olivieri2001towing}
Olivieri, A., Pistani, F., Avanzini, A., Stern, F., and Penna, R., 2001.
\newblock Towing tank experiments of resistance, sinkage and trim, boundary
  layer, wake, and free surface flow around a naval combatant insean 2340
  model.
\newblock Tech. rep., Iowa Univ Iowa City Coll of Engineering.

\bibitem{savitsky1964hydro}
Savitsky, D., 1964.
\newblock ``{Hydrodynamic Design of Planing Hulls}''.
\newblock {\em Marine Technology and SNAME News, \textbf{ 1}}(04), 10,
  pp.~71--95.

\bibitem{hollenbach1998estimating}
Hollenbach, K.~U., 1998.
\newblock ``Estimating resistance and propulsion for single-screw and
  twin-screw ships-ship technology research 45 (1998)''.
\newblock {\em Schiffstechnik, \textbf{ 45}}(2), p.~72.

\bibitem{hollenbach2007efficient}
Hollenbach, U., and Friesch, J., 2007.
\newblock ``Efficient hull forms--what can be gained''.
\newblock In Proceedings of the 1st International Conference on Ship
  Efficiency, Hamburg, Germany, pp.~8--9.

\bibitem{hart2010IMDO}
Hart, C.~G., and Vlahopoulos, N., 2010.
\newblock ``An integrated multidisciplinary particle swarm optimization
  approach to conceptual ship design''.
\newblock {\em Structural and Multidisciplinary Optimization, \textbf{ 41}},
  pp.~481--494.

\bibitem{diez2010robust}
Diez, M., and Peri, D., 2010.
\newblock ``Robust optimization for ship conceptual design''.
\newblock {\em Ocean Engineering, \textbf{ 37}}(11-12), pp.~966--977.

\bibitem{HybridAgent}
Daniels, A., and Parsons, M., 2008.
\newblock ``A hybrid agent — genetic algorithm approach to general
  arrangements''.
\newblock {\em Ship Technology Research, \textbf{ 55}}(2), pp.~78--86.

\bibitem{michell1898Wave}
Michell, J.~H., 1898.
\newblock ``Xi. the wave-resistance of a ship''.
\newblock {\em The London, Edinburgh, and Dublin Philosophical Magazine and
  Journal of Science, \textbf{ 45}}(272), pp.~106--123.

\bibitem{tuck1989wave}
Tuck, E.~O., 1989.
\newblock ``The wave resistance formula of jh michell (1898) and its
  significance to recent research in ship hydrodynamics''.
\newblock {\em The ANZIAM Journal, \textbf{ 30}}(4), pp.~365--377.

\bibitem{mantzaris1998rankine}
Mantzaris, D.~A., 1998.
\newblock ``A rankine panel method as a tool for the hydrodynamic design of
  complex marine vehicles''.
\newblock PhD thesis, Massachusetts Institute of Technology.

\bibitem{dawson1977practical}
Dawson, C., 1977.
\newblock ``A practical computer method for solving ship-wave problems''.
\newblock In Proceedings of Second International Conference on Numerical Ship
  Hydrodynamics, pp.~30--38.

\bibitem{supercloud}
Reuther, A., Kepner, J., Byun, C., Samsi, S., Arcand, W., Bestor, D., Bergeron,
  B., Gadepally, V., Houle, M., Hubbell, M., Jones, M., Klein, A., Milechin,
  L., Mullen, J., Prout, A., Rosa, A., Yee, C., and Michaleas, P., 2018.
\newblock ``Interactive supercomputing on 40,000 cores for machine learning and
  data analysis''.
\newblock In 2018 IEEE High Performance extreme Computing Conference (HPEC),
  pp.~1--6.

\end{thebibliography}

\end{document}